\documentclass[letterpaper, 10 pt, conference]{ieeeconf}  

\IEEEoverridecommandlockouts                              

\usepackage{graphics} 
\usepackage{epsfig} 
\usepackage{mathptmx} 
\usepackage{times} 
\usepackage{amsmath} 
\usepackage{amssymb}  

\usepackage{graphicx}
\usepackage{subcaption}
\usepackage{gensymb}
\usepackage{algorithm}
\usepackage[noend]{algpseudocode}
\usepackage{array}
\usepackage{multirow}
\usepackage{multicol}
\usepackage[normalem]{ulem}
\usepackage{xcolor}

\newcommand{\spacespace}[1]{\chi_{#1}}
\newcommand{\pathset}{\varrho}
\newcommand{\spacecone}[1]{K({#1}, d_{s_{max}})}

\title{\LARGE \bf
LSwarm: Efficient Collision Avoidance for Large Swarms with Coverage Constraints in Complex Urban Scenes
}

\author{Senthil Hariharan Arul, Adarsh Jagan Sathyamoorthy, Shivang Patel,\\ Michael Otte, Huan Xu, Ming C Lin, and Dinesh Manocha
}

\begin{document}

\maketitle
\thispagestyle{empty}
\pagestyle{empty}

\begin{abstract}
In this paper, we address the problem of collision avoidance for a swarm of UAVs used for continuous surveillance of an urban environment. Our method, LSwarm, efficiently avoids collisions with static obstacles, dynamic obstacles and other agents in 3-D urban environments while considering coverage constraints. LSwarm computes collision avoiding velocities that (i) maximize the conformity of an agent to an optimal path given by a global coverage strategy and (ii) ensure sufficient resolution of the coverage data collected by each agent. Our algorithm is formulated based on ORCA (Optimal Reciprocal Collision Avoidance) and is scalable with respect to the size of the swarm. We evaluate the coverage performance of LSwarm in realistic simulations of a swarm of quadrotors in complex urban models. In practice, our approach can generate global trajectories in a few seconds and can compute collision avoiding velocities for a swarm composed of tens to hundreds of agents in a few milliseconds on dense urban scenes consisting of tens of buildings.
\end{abstract}

\section{INTRODUCTION}

Recent advances in multi-rotor UAVs have created new areas of application, including surveillance, search and rescue, and monitoring. Continuous surveillance involves gathering sensory data (such as from a camera) from all regions in an area or volume of interest by traversing a path optimized for maximum coverage while accounting for static and dynamic obstacles \cite{c9}. For applications like surveillance, using a single UAV/agent may not be effective, when time, battery capacity, reliability, and coverage performance are considered.

An obvious solution to this problem is to use a swarm of agents and have each agent follow an optimal path provided by a global coverage strategy that maximizes the covered area or the information gathered from the environment. We use ``optimal coverage path" or ``global coverage path" to refer to this global path in this paper. Many techniques have been proposed that have formulated this optimal coverage path for multi-agent systems \cite{c13, c20, c21, c25, c30}. 

However, some of the existing swarm algorithms ignore collisions between the swarm agents and dynamic obstacles in the environment~\cite{c13, c21}, while others do not provide solutions that can be scaled to hundreds of agents. In addition, making an agent hover to avoid collision would not work when multiple dynamic obstacles approach head-on. When an agent encounters large numbers of dynamic obstacles, a haphazard maneuver to avoid collisions will again lead to loss of valuable data on the ground. In addition, the resolution of the data gathered is important to ensure its usefulness and it is an important factor to consider while performing collision avoidance maneuvers.




   \begin{figure}[t]
      \centering
      \includegraphics[height=2.0in,width=3.0in]{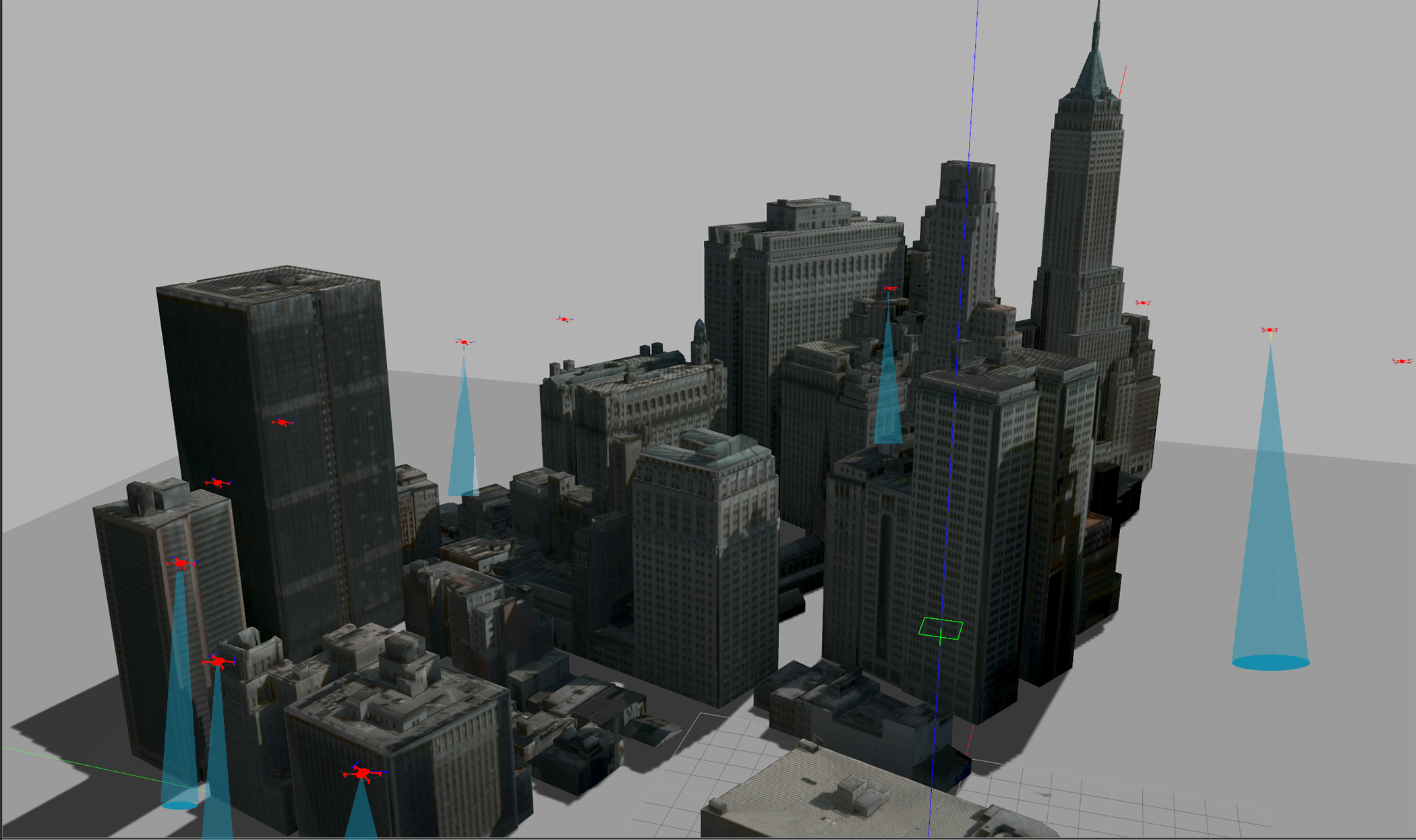}
      \caption {Simulation of $30$ quadrotors surveying our ``High-Dense" urban environment using LSwarm for collision avoidance with agents, dynamic obstacles and static obstacles while considering constraints on coverage.}
      \label{fig:highdense}
   \end{figure}
   
{\bf Objective:} 
In this paper, we address the complex problem of collision avoidance in a swarm of quadrotors during continuous surveillance of an urban environment. We impose constraints on coverage area and data resolution such that the quadrotors lose minimal ground coverage information while deviating to avoid collisions. 

{\bf Main Contributions:} 
We present LSwarm, a local collision avoidance method for quadrotor swarms performing continuous surveillance of large, complex 3-D urban environment (see Fig.\ref{fig:highdense}). LSwarm builds on ORCA \cite{c26} (a Velocity Obstacle based collision avoidance method), and to the best of our knowledge, it is the first implementation that considers all of the following:

\textbf{(a)} \textit{Collision avoidance with agents \& dynamic obstacles,}

\textbf{(b)} \textit{Collision avoidance with static obstacles,}

\textbf{(c)} \textit{Scalability to large number of agents,} 

\textbf{(d)} \textit{Acceleration Limits (Dynamics constraint),}

\textbf{(e)} \textit{Constraints on coverage,} 

\textbf{(f)} \textit{Uncertainty in position and velocity.} 

Our coverage constraints consider the following:

{\bf (i)} Each agent conforms maximally to an optimal path given by a global coverage strategy during collision avoidance. In other words, the calculated collision avoidance velocities for each agent also minimizes the coverage area loss (defined in Section III.A). {\bf (ii)} The coverage data obtained by the agents' sensors have sufficient resolution (defined by Ground Sampling Distance (GSD)).

LSwarm employs a precomputed look-up table containing coverage area overlaps for various collision avoiding velocities (Section IV.G) so that collision avoidance under coverage constraints can be executed in real-time. Our collision avoidance method is compatible with any global coverage strategy that provides waypoints. We use a simple lawn mower strategy to provide the global coverage path in this paper.

We evaluate LSwarm's performance with respect to ORCA in urban environments which contain a plethora of static obstacles such as buildings of various heights, and dynamic obstacles in the form of birds,  helicopters and other swarm agents. We perform accurate simulations of the quadrotor swarm in four different 3-D urban scenes. Depending on the model, the number of buildings vary between $10$ to $80$, and the height of buildings vary between $5$ to $30$ meters. Our algorithm takes  milliseconds to compute collision-free velocities for 50 quadrotor agents on eight cores. We observe that LSwarm provides an improvement of up to a factor of two in coverage area over the use of ORCA, as the number of dynamic obstacles increases.

We first survey the previous work in coverage strategies and collision avoidance in Section II. We formally define our coverage constraints and give a description of the global lawnmower strategy in Section III, and present our local collision avoidance algorithm in Section IV. Section V presents the experiments and results of our implementation. Section VI summarizes the paper and suggests possible directions for future work.
\section{PREVIOUS WORK}
Swarm agents such as UAVs have been an active area of research, due to their flexibility and a wide range of applicability in surveillance, search and rescue operations, and other information-gathering scenarios. A recent survey on control, planning, and design of aerial swarm robots can be found in \cite{c4}.

\subsection{Swarm Control Strategies for Coverage Optimization}
In this section, we give a brief overview of prior work on swarm-based coverage optimization, and collision avoidance strategies. 

{\bf Coverage Optimization:} 
Urban area surveillance with added localization constraints during trajectory generation and optimization of the final swarm distribution to cover high priority surveillance regions is presented in \cite{c21}. Julian et al. present a scalable consensus-based approximation method to generate trajectories to maximize the information gathered by a swarm in \cite{c13}. 
{ Dames et al. \cite{c33}, introduce a new formulation to track an unknown number of mobile targets using a team of robots based on random finite sets. Their method employs the Probability Hypothesis Density (PHD) filter to simultaneously estimate the number of targets and their positions after which the robots greedily choose actions to maximize certain control objectives. In \cite{c34} a distributed formulation of the PHD filter is used to search for and track the targets, which is combined with Lloyd's algorithm to control the robots' motions.}
Velagapudi et al. \cite{c30} present a distributed planner to explore moderately constrained environments that also accounts for communication constraints in large swarms in 2-D. { A receding horizon planning strategy for multi-robot coverage in partially known environments is presented by Das et al. in \cite{rhocop}.}

{\bf Collision Avoidance through precomputed trajectories:}\\
Saha et al. \cite{c20} demonstrate a scalable incremental motion planning method to find the order in which the robots should be dispatched to their destination in constricted environments.  Although this approach is useful for planning in narrow spaces in an urban setting, the time taken to precompute trajectories is high for even a group of tens of robots. Turpin et al. \cite{c25} present a centralized trajectory generation algorithm with optimality guarantees along with a scalable decentralized case. However, they assume the absence of static obstacles. A centralized feedback control law for uniform and ergodic domain coverage using multi-agent systems is shown in \cite{c18}. Trajectory generation for hundreds of quadrotors in dense continuous environments is demonstrated in \cite{c12}. This approach generates an inter-robot conflict annotated roadmap, then solves a generalized Multi-Agent Path Finding (MAPF) problem to avoid conflicts in the roadmap, while continuously smoothening the trajectory. 
A parallelization formulation for centralized trajectory generation methods for swarms is shown in \cite{c10}.

In contrast to previous work, our method combines a global coverage strategy (based on the lawnmower strategy) with a local collision avoidance scheme (based on Velocity Obstacle) which avoids collisions on the fly. The global coverage strategy in our case accounts only for avoiding static obstacles and not the inter-agent collisions in the swarm. { The decentralized local collision avoidance formulation can be scaled up to a large number of swarm agents, avoids having a single point of failure and also accounts for the dynamic nature of the environment such as dealing with obstacles not known a priori.}

\subsection{Space Filling Curves}
Multiple works have investigated applying and simplifying the implementation of space-filling curves such as the lawnmower and Hilbert curves for global coverage. 
The first formal work on sweep search that used a probabilistic lawnmower sweep method is presented in \cite{c15}. In \cite{c3}, the boustrophedon cellular decomposition of the free space is discussed. This work drastically simplifies the number of cellular decompositions compared to previous methods and allows a robot to cover each cell with back and forth lawnmower motions. Kong et al.~\cite{c14} build on this work by allocating multiple cooperating and communicating robots to cover different areas in the environment in a distributed fashion. Vincent and Rubin~\cite{c31} present a cooperative search using a swarm of UAVs that follows lawnmower trajectories to detect an evading target. The use of Hilbert space-filling curves for a geographic search using robot swarms is explored in \cite{c24}.   Our method adopts the lawnmower strategy \cite{c3} to plan for optimal coverage paths, which are then used as a constraint for performing local collision avoidance that minimizes deviation from the optimal coverage using a team of aerial robots.

\subsection {VO-based collision avoidance}
Collision avoidance with dynamic obstacles has been investigated extensively in \cite{c6,c8,c11,c17}. These approaches estimate the future positions of obstacles using their observed velocities. Velocity Obstacle (VO) \cite{c7} methods transform the collision avoidance problem in the workspace to an equivalent formulation in the velocity space. Reciprocal Velocity Obstacles (RVO) \cite{c27} improved on VO to tackle robot-robot collision avoidance, where each agent chooses its own velocity based on the velocity of other agents. Optimal Reciprocal Collision Avoidance (ORCA) \cite{c26} further built upon RVO by providing a sufficient condition for multiple robots to avoid collisions among one another, thus guaranteeing collision-free navigation.  ORCA was extended to 3-D in \cite{c29} and to include a proximity based static obstacle avoidance in~\cite{c1}. { Proximal Policy Optimization (PPO), a policy gradient based reinforcement learning algorithm for decentralized collision avoidance is shown in \cite{ppo}. Agents that use PPO have slightly better collision avoidance success rates, reach their goals sooner and have higher average speeds.} Breitenmoser and Martinoli~\cite{c2} combine multi-robot coverage and reciprocal collision avoidance and evaluate the new method's benefits in several scenarios.

{ Although ORCA has several advantages since it is decentralized, and scalable, the algorithm assumes perfect state information in terms of position and velocity of each agent and does not model uncertainty.}

\subsection{{ Modeling Uncertainties}}
{ Many techniques have been proposed that extend the VO based algorithms to be more practical by considering the uncertainty in localization of the agents and sensing of nearby obstacles. Hennes et al. \cite{c39} provide an Adaptive Monte Carlo localization-based solution (AMCL) for modeling the pose of an agent and basic communication between agents to share the positions and velocities of nearby agents. Although this method works well for ground robots (2-D scenarios), the dependence of AMCL on odometry and laser data makes it hard to use it for UAVs. Recent advances in sensor technologies such as GPS-RTK (Real Time Kinematics) have improved the localization accuracy of the agents to centimeter levels. Thus, communicating such information between nearby swarm agents could result in a better collision avoidance scheme.
 Even in GPS-denied environments, works such as PIXHAWK \cite{c38} have shown the feasibility of vision-based localization and the use of stereo cameras to detect nearby obstacles with all computations being performed on an on-board computer. Such techniques can be used to detect and avoid unreactive dynamic obstacles such as birds.  Other techniques take into account sensing and actuation uncertainties. Gopalakrishnan et al. \cite{c37} address this problem by presenting a tractable approximation of chance constraints along with the velocity constraints of RVO for 2D agents \cite{c27}. Snape et al. \cite{c35} and Kamel et al. \cite{c36} present Kalman filtering based methods to vary the safe distance between the agents based on position and velocity uncertainties. We make use of a similar Kalman filtering based method to handle uncertainties among agents and obstacles in LSwarm.   
}

{\bf Simulation:} OpenUAV, the state of the art in UAV simulation is presented in \cite{c22}. It is a cloud-enabled testbed for UAVs using ROS, Gazebo and PX4 simulation packages. { The current implementation of OpenUAV is capable of simulating 9 UAVs with cameras and around 30 UAVs without simulated cameras. This number can improve as OpenUAV moves to a more scalable cloud computing platform.}

While constraints to avoid collisions between swarm members and static obstacles are included in most trajectory generation methods, issues like the absence of dynamic obstacle avoidance, ability to adapt to changing scenes, and issues in scalability are still prevalent in most of these methods. ORCA is insufficient for collision avoidance in urban scenarios since it does not have a suitable static collision avoidance method for dense 3-D urban environments. LSwarm overcomes such limitations of ORCA and retains its advantages such as scalability and compatibility with global planners.
\section{SWARM COVERAGE IN URBAN SCENES}

In this section, we formally define our constraints on coverage loss and resolution of the collected data. We state the assumptions we make for each swarm agent and then describe the global lawnmower strategy that provides the optimal coverage path. LSwarm is compatible with any global method that generates waypoints { for each agent}, and in this paper we use a lawnmower strategy due to its simplicity.

\subsection{Coverage Constraint Definition}
As stated earlier, LSwarm's coverage constraints ensure that collision avoidance velocities are calculated such that coverage area loss is minimized and the resolution of collected data is satisfactory. 

\subsubsection{Coverage Area Loss}
Each swarm agent's primary objective is to follow the optimal coverage path as closely as possible. The optimal coverage path provides the maximum coverage of the environment, and any deviation from this path would result in loss of coverage area. The following definitions apply to each quadrotor in the swarm.

Let $area_{pref}$ be the \textit{preferred} area that would have been covered if a quadrotor followed the optimal coverage path perfectly for a given period of time. Let $area_{actual}$ be the \textit{actual} area covered by the quadrotor. When the quadrotor does not face any obstacle while on the optimal coverage path, $area_{actual} = area_{pref}$. In cases where the quadrotor faces obstacles, $area_{actual}\bigcap area_{pref} < area_{pref}$. Let $area_{overlap}$ denote the overlap area between $area_{pref}$ and $area_{actual}$. Then,
\begin{equation*}
    area_{overlap} = area_{pref} \bigcap area_{actual}
\end{equation*}

The overlap ratio is given by,
\begin{equation*}
    overlap\hspace{0.5em}ratio = area_{overlap} / area_{pref}
\end{equation*}

The coverage loss is then defined as $(1 - overlap\hspace{0.5em}ratio)$. When $overlap\hspace{0.5em}ratio = 1$, (optimal path is perfectly followed), the coverage loss is 0.\\

\subsubsection{Sensor Resolution}
We assume that each quadrotor or the agent in our swarm has a sensor (a camera in our case) to gather information from the environment. The sensor is modeled to have a conical view of the environment with a fixed apex angle. The radius $r$ of the base circle of this view-cone would depend on the altitude $h$ at which the quadrotor is flying as:

\begin{equation}\label{htan}
  r = h \cdot \tan{\theta}
\end{equation}
where $\theta$ is half of the apex angle of the cone.

This means that the sensor covers more area when the quadrotor is flying at a higher altitude, but the resolution of the sensor output at this higher altitude may not be satisfactory. In some collision avoidance scenarios, the quadrotor might change its altitude which again could result in loss of resolution. 

\begin{figure}[t]
      \centering
      \includegraphics[height=2.75in, width=2.75in]{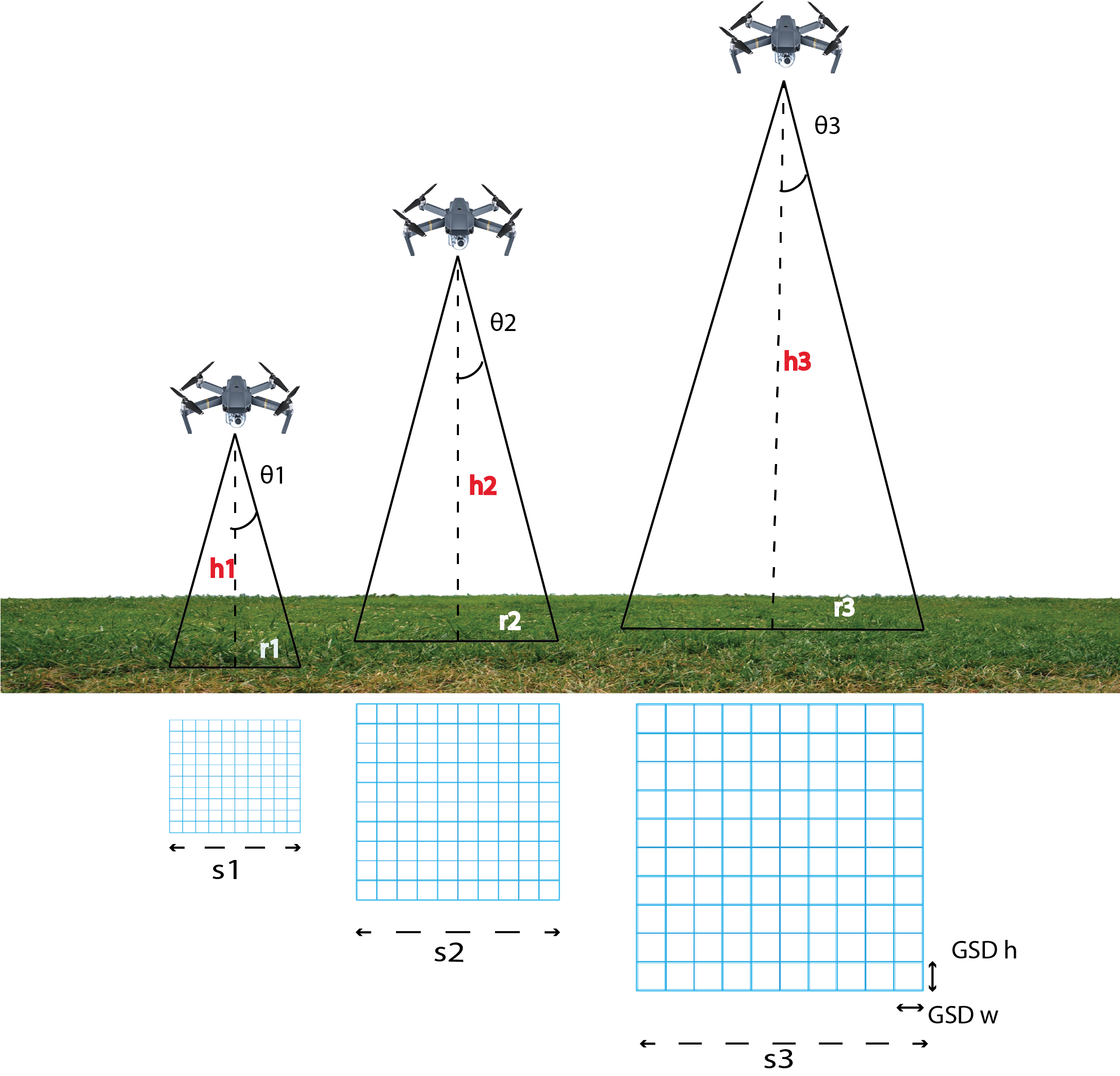}
      \caption{Change in area of the coverage grid for different flying altitudes of the quadrotor is shown. $GSD_h$ and $GSD_w$ are a measure of the height and width of the pixels in the coverage grid. As the grid becomes larger, the finer details on the ground are missed.}
      \label{fig2}
   \end{figure}

To ensure that there are no detrimental effects to resolution, we use a measure called Ground Sampling Distance (GSD) and use it in our coverage constraints. GSD is the length on the ground denoted by one pixel on the image given by our camera. GSD increases with altitude and the greater the value of GSD, the poorer the resolution (see Fig.\ref{fig2}). GSD is chosen as a metric for resolution because it provides a linear relationship between the flying altitude of the quadrotor and the features that can easily be covered on the ground from that altitude. For example, if our objective is to view the people on the ground, we can calculate the optimal height at which the quadrotor has to fly to make at least two pixels represent a person. Since GSD is calculated for a rectangular image, we take the largest area square within the base circle of the sensor cone for our calculation of GSD. The area of this square (of side s) is considered to be the area covered by a quadrotor at one instant of time.  The GSD for a rectangular image is generally calculated as,
\begin{equation}
\begin{split}
    GSD_{h} = K_{h}\cdot h \\
    GSD_{w} = K_{w}\cdot h \\
\end{split}
\end{equation}

\noindent where h is the flying altitude and K is a constant calculated as:
\begin{equation}
\begin{split}
    K_{h} = \frac{Sensor\,Height}{f \cdot Image\,Height} \\
    K_{w} = \frac{Sensor\,width}{f \cdot Image\,Width}
\end{split}
\end{equation}

Where the sensor height and sensor width are the dimensions of the camera lens and \textit{f} is the focal length of the lens. {Since we are considering a square camera footprint, {$GSD_{h}$} is equal to {$GSD_{w}$}.}

The relationship between the side of the largest area base square and the altitude from $\left(\ref{htan}\right)$ is given as:
\begin{equation}\label{Eqn.9}
    s = \frac{h\cdot\tan{\theta}}{\sqrt{2}}
\end{equation}

\subsection{Assumptions on Swarm Agents}
We assume that the sensor attached to the agent always faces downwards irrespective of the orientation of the agent. This can be achieved by attaching the sensor to a gimbal. We also assume that each agent knows its own position and velocity, and can sense the positions and velocities of other agents around it. All of these values could have some level of uncertainty in them. Section IV.E discusses our Kalman filtering method to handle this uncertainty.

\subsection{Global Coverage Strategy}
 We use a simple lawnmower sweep to generate our optimal coverage path.
Lawn mower sweep, which is also known as Boustrophedon coverage, is a simple yet widely used strategy that guarantees 100\% coverage of a search space \cite{c3, c19}.
Some implementations \cite{c3} segregate the obstacle and the non-obstacle region and perform a separate search sweep in each connected non-obstacle region.
Similar ideas have also been used in unknown terrain; for example, \cite{c19} uses a combination of boustrophedon and A* to, respectively, (a) perform a search sweep in unknown terrain, and (b) optimally traverse the known terrain. 

Using lawn mower sweep in urban environments provides a unique set of challenges. Covering occluded areas where buildings are densely constructed is one challenge that is not typically addressed by the standard two-dimensional implementations of the approach. The occlusion problem can be mitigated by having agents adjust their height to fly over the buildings that create such occlusion. In other words, connections between different regions of ground-level search are possible by having agents perform a temporary change in altitude by flying over buildings. 

   \begin{figure}[t]
      \centering
      \includegraphics[height=2in, width=3.25in]{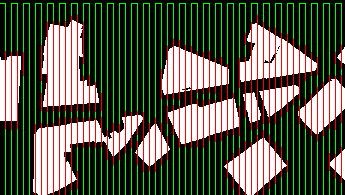}
      \caption{Top-down view of Lawnmower waypoints over a city block. The white part represents the buildings (obstacles) and the black regions represent obstacle-free space. A part of lawn mower sweep is shown in green and red colors. Green color represents the optimal height of operation while the red color represents an elevated path to avoid buildings.}
      \label{fig3}
   \end{figure}

\subsubsection{Lawn mower sweep: formalization and problem statement}
This subsection will formulate Lawn Mower algorithm mathematically.

Given, a space $\spacespace{}$. Defining $\spacespace{obs}$ and $\spacespace{free}$, as obstacle space and free space respectively, such that $\spacespace{obs}$, $\spacespace{free}$ $\subset$ $\spacespace{}$. A path $\rho$ is a function $\rho : [0,1] \rightarrow \spacespace{free}$ which maps the path to a point in the free space. Therefore we can define

\begin{equation}\label{Eqn.3-3}
    \pathset = \bigcup_{i =1}^{n} \{\rho_i\}
\end{equation}

\noindent
where $n$ is number of agents. Now let $\spacespace{search}$ be the search space such that $\spacespace{search} \subseteq \{\spacespace{free} \cap groundplane\}$. Assuming agent's camera sensor always facing downwards, the volume of space that is sensed by the sensor forms a conical shape below the agent. Let $\spacecone{X}$ is the set of points in a downward facing camera at $X$ and extended below $X$ till $d_{s_{max}}$ forming a 3D conical space. Here $d_{s_{max}}$ might be inferred as the valid sensing distance.
The multi-robot coverage problem is formally defined as follows:

\textbf{Multi-robot coverage problem:}\\ 
Find $\varrho$ such that $\spacespace{search} \backslash \spacespace{swept} = \emptyset$, where $\spacespace{swept}$ is given by:
\begin{equation}\label{Eqn.3-4}
    \spacespace{swept} = \{ X \mid X \in \bigcup_{\hat{x} \in \varrho} \spacecone{\hat{x}} \}
\end{equation}
\noindent
and where $\spacespace{search} \subseteq \spacespace{swept}$.

\subsubsection{Global Solution}
Our global lawnmower solution experiments take the quadrotors' sensor model and a resolution measure for the sensor output into account and precompute the waypoints (as seen in Fig.\ref{fig3}) over the environment. We use a simple approach that extends a standard lawn mower sweep from two-dimensions to three dimensions by determining the optimal flying altitude to obtain a good resolution in the output and the necessary altitude changes to avoid collisions with building along the sweep path, and then augmenting the height components of the path accordingly. It is possible to use more sophisticated methods, which calculate a sweep path that minimizes the number of altitude changes required over the entire search. We choose a  simple method so that we can easily compute such waypoints for complex urban scenes with a large number of buildings. 

\section{LOCAL COLLISION AVOIDANCE WITH COVERAGE CONSTRAINTS}
In this section, we provide a brief introduction to ORCA and then define our local collision avoidance method with coverage constraints.  

\subsection{Symbols and Notations}
A list of symbols and their definitions used in the formulation of ORCA \cite{c26} is shown in Table \ref{tab:Table 1}. A brief explanation for ORCA using these symbols is provided in Section IV.B.

\begin{table}
    \begin{tabular}{|c|p{60mm}|}
      \hline
      \textbf{Symbol} & \textbf{Meaning} \\
      \hline 
      \rule{0pt}{12pt} $v_{A}^{pref}$ & Velocity of Agent A directed towards goal position in the absence of obstacles \\
      \hline
      
      \rule{0pt}{12pt} $v_{A}^{new}$ & New Velocity chosen by Agent A \\
      \hline
     
      \rule{0pt}{12pt} $v_{A}^{opt}$ & Optimal velocity for agent A \\
      \hline
     
      \rule{0pt}{12pt} $VO_{A|B}^{\tau}$ & Velocity Obstacle induced by B on A for time $\tau$ \\
      \hline
     
      \rule{0pt}{12pt} $D \left( \textbf{p}, r \right)$ & Disc centered at \textbf{p} and radius r \\
      \hline
     
      \rule{0pt}{12pt} $\textbf{X} \oplus \textbf{Y}$ & Minkowski sum of sets \textbf{X} and \textbf{Y}\\
      \hline
      
      \rule{0pt}{12pt} ${P_A}$ & Current Position of Agent A\\
      \hline 

      \rule{0pt}{12pt} ${WP_A}^{Next}$ & Next Waypoint for Agent A\\
      \hline      
      
      \rule{0pt}{12pt} ${WP_A}^{Prev}$ & Previous Waypoint for Agent A\\
      \hline        

      \rule{0pt}{12pt} $CP_A$ & Closest Point on line segment ${WP_A}^{Next}$ ${WP_A}^{Prev}$ from current position of Agent A\\
      \hline       
     
      $ORCA_{A|B}^{\tau}$ & Optimized collision avoiding velocity set for A induced by B during time $\tau$ \\
      \hline
    \end{tabular}
    \caption{\label{tab:Table 1} Definition of Symbols used in ORCA formulation}
\end{table}

\subsection{Optimal Reciprocal Collision Avoidance}

\subsubsection{\textbf{Velocity Obstacles}}
Consider a pair of agents A and B. The velocity obstacle (VO) for the pair is defined as the set of all relative velocities of A with respect to B that will result in a collision between A and B before time $\tau$. It is formally defined as,
\begin{equation}\label{Eqn.1}
 VO_{A|B}^{\tau} = \left\{ \textbf{v} | \exists t \in [0,\tau]::t\textbf{v} \in D \left(\textbf{p}_{B} - \textbf{p}_{A} , r_{A} + r_{B} \right) \right\} 
\end{equation}

\noindent where $D \left( \textbf{p}, r \right)$ is a disk centered at \textbf{p} with radius r which is defined as: 
\begin{equation}\label{Eqn.2}
 D \left( \textbf{p}, r \right) = \left\{ \textbf{q} \, | \, ||\textbf{q} - \textbf{p}|| < r \right\}. 
\end{equation}

In 2-D, ORCA represents all agents as discs which is extended to a sphere representation in 3-D. The velocity obstacle for such a case can be geometrically interpreted as a truncated cone with its apex at the origin of the velocity space and its legs tangent to the disc of radius $r_{A}$ + $r_{B}$, centered at $\textbf{p}_{B} - \textbf{p}_{A}$. We can infer from the formulation of VO that for any velocity $\textbf{v}_{B} \in V_{B}$ , if $\textbf{v}_{A} \not\in VO_{A|B}^{\tau}\oplus V_{B}$, then A and B are guaranteed to avoid collision for at least time $\tau$. The ORCA algorithm calculates optimal collision avoidance velocity sets from which new $\textbf{v}_{A}$ and $\textbf{v}_{B}$ can be chosen.

\subsubsection{\textbf{Optimizing the Velocity Computation}}
ORCA tries to optimize the new chosen velocities to be close to the agents' optimal velocities. An agent's optimal velocity can be one of three values: zero, preferred velocity $v_{A}^{pref}$ or the agent's current velocity. The objective is to calculate the velocity sets $ORCA_{A|B}^{\tau}$ for A and $ORCA_{B|A}^{\tau}$ for B. These sets contain more velocities close to $v_{A}^{opt}$ and $v_{B}^{opt}$, respectively, than any other pair of sets of reciprocally collision-avoiding velocities.

If suppose $v_{A}^{opt} - v_{B}^{opt} \in VO_{A|B}^{\tau}$ , then the two agents may collide before time $\tau$. To achieve collision avoidance with the least amount of effort, ORCA finds a relative velocity on the boundary of $VO_{A|B}^{\tau}$ that is closest to $v_{A}^{opt} - v_{B}^{opt}$. 


\textbf{Reactive Obstacles:}
ORCA lets each agent take half of the responsibility for changing their velocities such that their relative velocity lies outside $VO_{A|B}^{\tau}$. Based on this formulation, the ORCA velocity set for A is given as:
\begin{equation}\label{Eqn.5}
    ORCA_{A|B}^{\tau} = \left\{ v \, | \left(v - \left(v_{A}^{opt} + \frac{1}{2}u \right) \right) \cdot \textit{$\hat{n}$} \ge 0 \right\} ,
\end{equation}

\noindent where \textit{u} is the vector from $v_{A}^{opt} - v_{B}^{opt}$ to the VO boundary point and \textit{$\hat{n}$} is the outward normal at point $\left( v_{A}^{opt} - v_{B}^{opt}\right) + u$ on the boundary of $VO_{A|B}^{\tau}$. \\

\textbf{Non-reactive Obstacles:}
The dynamic obstacles that cannot change their velocities to avoid a collision are referred to as non-reactive dynamic obstacles. Since such obstacles take no responsibility to avoid a collision, the ORCA velocity set for an agent in such cases is given by:
\begin{equation}\label{rand}
    ORCA_{A|B}^{\tau} = \left\{ v \, | \left(v - (v_{A}^{opt} + u) \right) \cdot \textit{$\hat{n}$} \ge 0 \right\} ,
\end{equation}
The same expression also applies to static obstacles. \\

As an extension for a swarm of `n' agents, the permitted velocity set for agent A denoted as $ORCA_{A}^{\tau}$ is the intersection of all half-planes induced by all agent B's. It is defined as:

\begin{equation}\label{Eqn.6}
 ORCA_{A}^{\tau} = \bigcap_{B \neq A} ORCA_{A|B}^{\tau}    
\end{equation}

The new velocity that agent A chooses would be the one that is closest to its preferred velocity $v_{A}^{pref}$ from $ORCA_{A}^{\tau}$. 

\begin{equation}\label{Eqn.7}
 v_{A}^{new} =  \underset{v \in ORCA_{A}^{\tau}}{\operatorname{argmin}} ||\,v - v_{A}^{pref}||  
\end{equation}

The new velocity is computed using linear programming. ORCA is a simple algorithm with low computational overhead. Therefore, it can be run online on each agent and is scalable to a large number of swarm agents. With its potential to be implemented on a large swarm, ORCA needs to be augmented with additional constraints to handle the coverage loss that arises from agents encountering a large number of dynamic obstacles, and a suitable static obstacle avoidance scheme.

\subsection{Collision Avoidance with Static Obstacles}
 ORCA models static obstacles using line segments in 2-D and using planes in 3-D. Such a model is inefficient and time consuming when used in dense 3-D urban environments. The lawn mower strategy computes the global coverage path while accounting for the static obstacles in the environment. In dense urban scenarios, the quadrotors might be required to maneuver close to buildings and any deviation from the global path during dynamic obstacle avoidance may cause them to collide with the buildings (scenario represented in Fig. \ref{staticcollisionavoidance}). To prevent such collisions, and to overcome ORCA's limitation, our method accounts for the static obstacles in the ORCA formulation (described in section IV.B) using proximity queries as suggest in \cite{c1}. 
 
\begin{figure}[t]
      \centering
      \includegraphics[height=1.8in, width=3.25in]{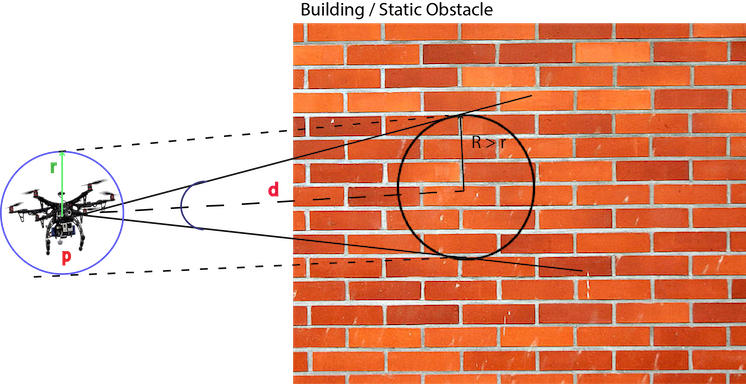}
      \caption{The Velocity Obstacle (truncated cone) constructed by a quadrotor when encountering static obstacles. The sphere with radius R is the set denoting the Minkowski sum of the quadrotor radius (\textit{r}) and the obstacle point radius ($\epsilon$)}
      \label{fig4}
   \end{figure}

\begin{figure}[t]
      \centering
      \includegraphics[height=1.8in, width=3.25in]{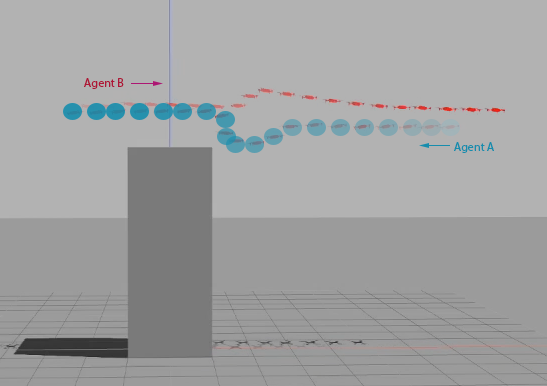}
      \includegraphics[height=1.8in, width=3.25in]{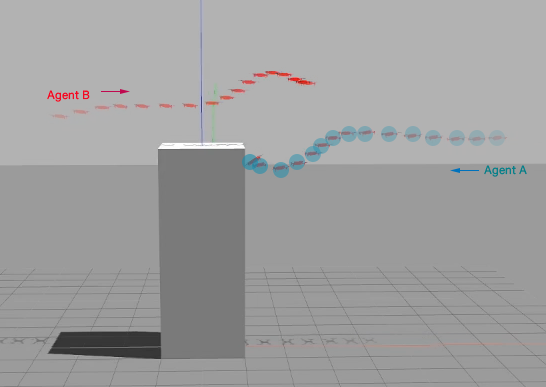}
      \caption{In the scenario above both agents have a collision free global path but dynamic obstacle avoidance deviates the path towards the building. (Top) LSwarm avoids the static obstacle while (Bottom) ORCA suffers Collision. Color intensity increases with time-step}
      \label{staticcollisionavoidance}
   \end{figure}

 Each quadrotor continuously monitors its surroundings and computes its proximity to static obstacles ($d$ in Fig.\ref{fig4}), like the buildings in an urban scene. The quadrotor's collision model is taken as a sphere with radius \textit{r}. The closest point on the static obstacle is considered as a point obstacle and the Minkowski sum is calculated considering a small positive value for the radius of the closest point $(\epsilon)$ and the quadrotor sphere's radius (\textit{r}). 
\begin{equation}
    R = r \oplus \epsilon
\end{equation}
We compute the Velocity Obstacle using the Minkowski sum as shown in Figure.4. We use the ORCA formulation for non-reactive obstacles as described in Section IV.B to select a new velocity for the agent that is closest to the its optimal velocity and lies outside the VO. Since this method considers only the closest obstacle point, it can be easily incorporated in a physical quadrotor system using simple depth sensors.\\

\noindent {\bf{Selecting Preferred Velocity:}} In ORCA, the agent's preferred velocity $v_{A}^{pref}$ is directed towards the next waypoint. In certain cases, with the deviation caused by collision avoidance, this $v_{A}^{pref}$ can be directed \textit{through} the buildings. The static obstacle avoidance described in Section IV.C will prevent collision with the buildings. However, since we only consider the closest points as obstacles, the agent may reach a deadlock trying to follow this preferred velocity to the next waypoint. To prevent such deadlocks, we update the $v_{A}^{pref}$ as being directed to the closest point along the line segment connecting the waypoint that the agent visited before the deviation, and next waypoint, the waypoint that the agent would have visited if not for the deviation.


\subsection{Dynamics Constraints}
ORCA assumes that agents can modify their velocities instantaneously. Hence, some fundamental dynamics constraints need to be considered before calculating new velocities for quadrotors. These dynamics constraints can be limits on the maximum velocity and acceleration of a quadrotor, which in turn translates to constraints on the roll, pitch and yaw angles. Quadrotors in the swarm may destabilize when there is a large change in their velocities, which results in large pitch or roll angles that might topple the quadrotors. To prevent such scenarios, we fix the maximum acceleration of the quadrotors and eliminate any velocity that does not obey the following constraint:
\begin{equation}
    v_{new} < v_{current} + a_{max} \cdot t
\end{equation}
This formulation helps keep the collision-free velocity space convex, thus ensuring fast computation time.
The search for appropriate velocities can be performed using acceleration velocity obstacles~\cite{c28}.

{ \subsection{Modeling Uncertainty}
We overcome ORCA's requirement for perfect sensing by using Kalman filtering to handle noise in the position and velocity of the swarm agents and obstacles. We use a Kalman filter to estimate the mean and covariance matrix of the positions and velocities of all moving entities. We then use the square root of the largest eigenvalue of an entity's covariance matrix as a measure to increase the radius of the bounding sphere of that entity. Thus, as the uncertainty in sensing increases, each agent takes a more conservative approach to avoid collisions. We observed that when noise was introduced in the state parameters of the agents and dynamic obstacles, the inclusion of Kalman filtering resulted in collision-free trajectories for the agents.}

\subsection{Dynamic Collision Avoidance with Coverage Constraints}
We first derive an objective function that needs to be maximized based on the overlap area that was described in section III.A. 

In the absence of dynamic obstacles, all quadrotors would fly between one waypoint to the next with their preferred velocities at the optimal altitude used by the lawnmower strategy. As mentioned previously, we consider that each quadrotor covers a square area at any instant of time. Therefore, the area covered by a quadrotor at any time { instant} $t \in [0,\tau]$ can be given as:
\begin{multline} \label{Eqn.10}
    area_{pref} = Area\left(x, y, s, t \right)\\ = \textrm{Area of square of side s centered at} \left(x, \, y\right)   
\end{multline}
where,  { $\tau$ is the time horizon over which we calculate the area, and}
\begin{align}
    x = v_{pref_x} \cdot  t\\ 
    y = v_{pref_y} \cdot  t\\
    s = \left(h - v_{pref_z}\cdot t\right) \cdot \frac{\tan{\theta}}{\sqrt{2}},
\end{align}
where $h$ is the initial height at which the quadrotor was flying at {t = 0.}\\

The area covered over a time horizon $\tau$ for which an agent's velocity is constant can be computed as:
\begin{equation}\label{Eqn.11}
    \bigcup_{\tau} Area\left(x, y, s, t \right)
\end{equation}
We sample the time horizon $\tau$ into smaller time steps t and forward simulate the velocity $v_{new}$ to calculate the area that would be covered if $v_{new}$ is chosen as the new velocity.

When dynamic obstacles are present, the quadrotors chooses a new velocity $v_{new}$ according to $\left( \ref{Eqn.7}\right)$, and $x, y$ and $s$ in equation $\left(\ref{Eqn.10}\right)$ would be with respect to $v_{new}$.

\begin{multline}\label{Eqn.14}
    area_{new} = \bigcup_{\tau} Area \bigg(v_{new_x} \cdot t, v_{new_y} \cdot  t,\\ \left(h - v_{new_z}\cdot  t\right) \cdot \frac{\tan{\theta}}{\sqrt{2}},  t \bigg)     
\end{multline}

The above equation is obtained as follows. The X and Y coordinates of the center of coverage squares are obtained from the product of the X and Y components of $v_{new}$ and t. The product of $v_{new_z}$ and  t gives us the change in the flying height or altitude of the quadrotor. 

\begin{align*}
    \Delta h = v_{new_z} \cdot t \\
    h_{new} = h - \Delta h = h - v_{new_z} \cdot t  \\
    s_{new} = h_{new} \cdot \frac{\tan{\theta}}{\sqrt{2}}
\end{align*}

The conformity to the global optimal coverage path automatically results in minimizing the loss of the coverage area. Conformity increases with increase in overlap between the areas in equations $\left( \ref{Eqn.10} \right)$ and $\left( \ref{Eqn.14}\right)$, while also limiting the GSD below an upper threshold.\\

Therefore, our objective function becomes:
\begin{multline}
    \max_{GSD \le GSD_{max}} \left(area_{overlap}\right) =\\ \max_{GSD \le GSD_{max}} \left(area_{pref} \bigcap area_{new}\right)
\end{multline}

\begin{figure}[t]
      \centering
      \includegraphics[height=2.0in, width=3.25in]{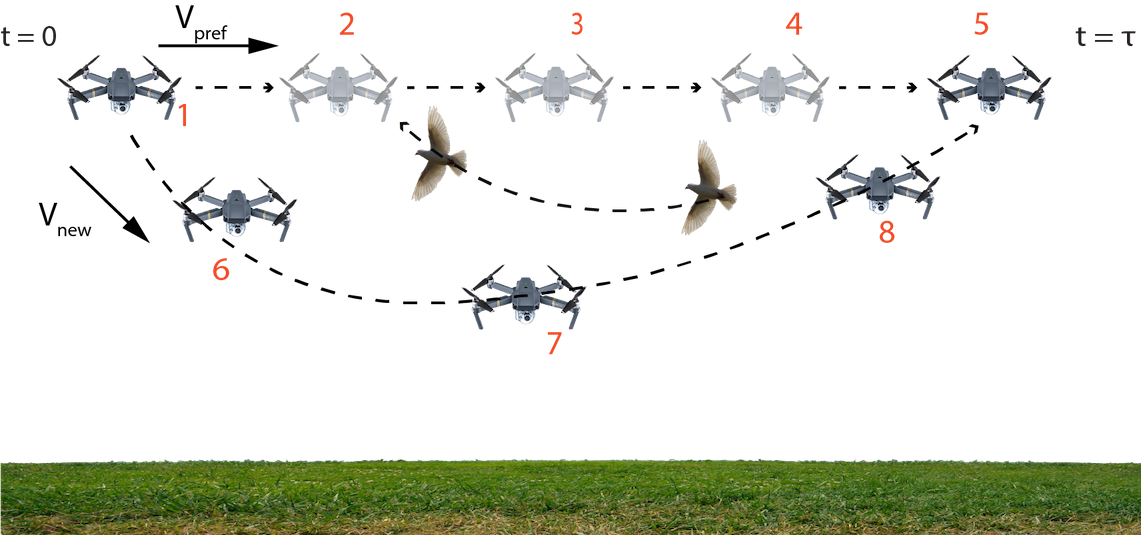}
      \caption{Quadrotor deviating downward when avoiding a dynamic obstacle (a bird). The translucent quadrotors represent the path that would have been taken in the absence of the dynamic obstacle, when v  = $v_{pref}$. The curve with the opaque quadrotors represents the path taken to avoid the obstacle when v  = $v_{new}$ and is computed by ORCA. The different states of the quadrotor are numbered. We highlight the coverage for these positions in Fig. \ref{overlap}.}
      \label{bird}
   \end{figure}

\begin{figure}[t]
      \centering
      \includegraphics[height=2.3in, width=3.5in]{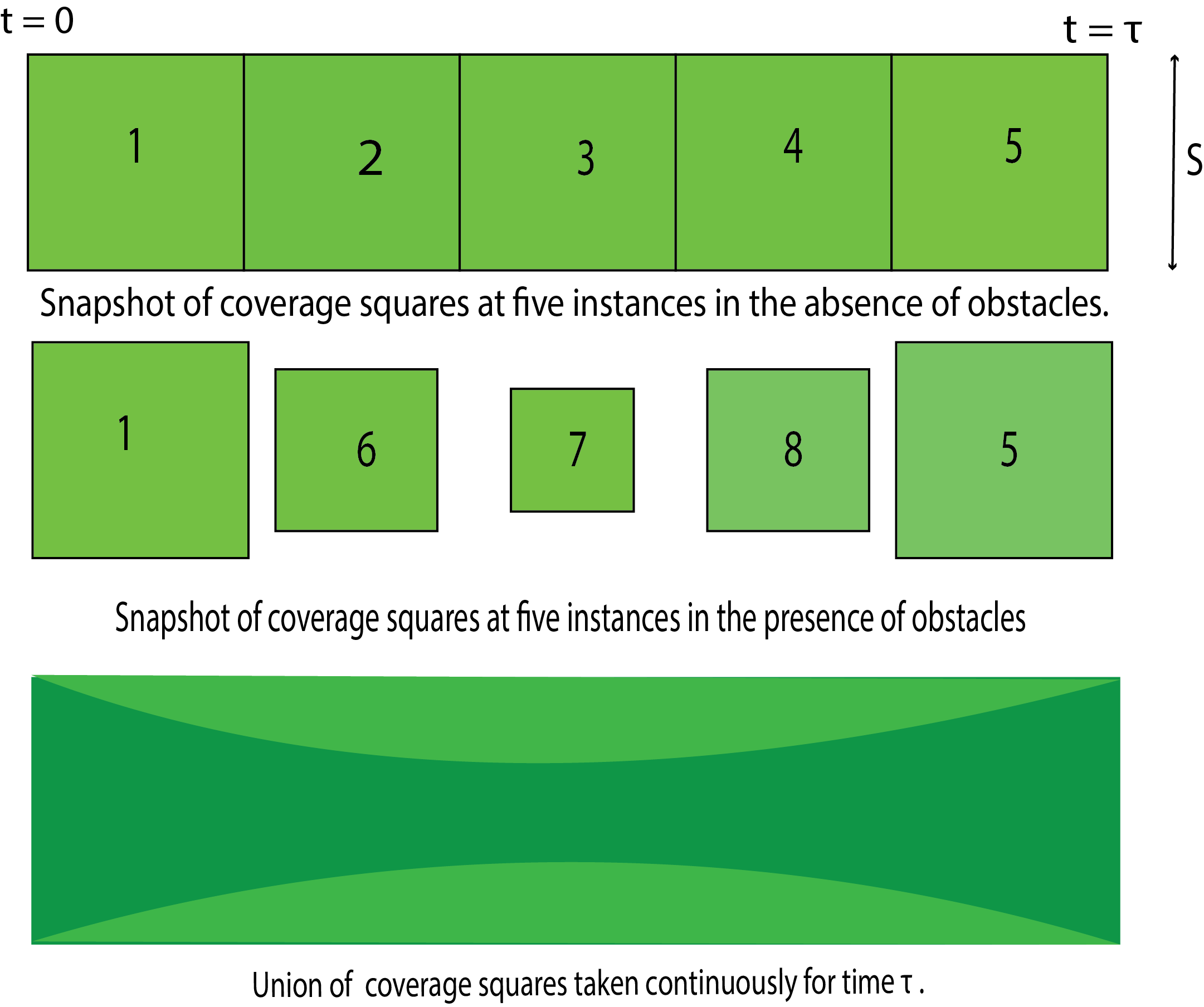}
      \caption{The coverage areas for the numbered stated in Fig. \ref{bird} for different locations of the agent. We observe lowest coverage at location 7 and maximum coverage at locations 1 and 5. [Top] The preferred coverage $area_{pref}$, when $v_{pref}$ is the velocity of the quadrotor. [Middle] The new coverage $area_{new}$, when $v_{new}$ is chosen as the velocity of the quadrotor. [Bottom] The overlap area (shown in dark green) between $area_{pref}$ and $area_{new}$ when computed continuously for time $\tau$. This overlap area need to be maximized.}
      \label{overlap}
   \end{figure}

A collision avoidance scenario is shown in Fig. \ref{bird}, and the corresponding coverage area for different states of the quadrotor are shown in Fig. \ref{overlap}.

Overlap area can be computed by methods such as clipping, but performing clipping in each time-step for each agent would be computationally expensive. 
We avoid the above issue using a look up table with pre-computed values for overlap area as explained in the next section.

\subsection{LSwarm Acceleration}
The formulation of $area_{overlap}$ as a function of $v_{new}$ and GSD is required for maximizing it for a given preferred velocity $v_{pref}$. Such a function depends on the square of the flying altitude and sinusoids of $v_{new}$ vector's angles with the Y and Z axes (we take X, Y and Z axes as the Roll, Pitch and Yaw axes respectively). The function cannot be solved for a $v_{new}$ that maximizes it in realtime using simple optimization algorithms (e.g., linear programming). Therefore, we use a Look Up Table (LUT) that contains precomputed values for the overlap area for various new velocities. At runtime, we select a new velocity based on its corresponding overlap area from this LUT and on certain conditions. The construction of the LUT and velocity selection are discussed in the following sections.  

\subsubsection{Constructing the Look Up Table}
Let us consider the unit vector $\left(1, 0, 0 \right)$ and call it $v_{unit}$. This unit vector corresponds to the forward direction for the quadrotor in the quadrotor's frame of reference and can be transformed to any preferred velocity vector by using { the standard 3x3 rotational transformation matrix \textbf{(R)}} as,
\begin{equation}
    v_{pref} = R_{trans} * v_{unit}
\end{equation}
For a time horizon $\tau$, we calculate $area_{unit}$ corresponding to $v_{unit}$ as given in (\ref{Eqn.10}) {assuming a flying altitude of 5m}. We rotate $v_{unit}$ (about Y and Z axes) in all possible forward directions to evaluate how such rotations affect the overlap area with $area_{unit}$. 

We first apply a rotation to $v_{unit}$ about the Z-axis as,
\begin{equation}
    v_{\alpha} = R_{Z\alpha} \cdot v_{unit},
\end{equation}
where $R_{Z \alpha}$ is the rotation matrix denoting a rotation by angle $\alpha$ about Z-axis which is given as: \\
\begin{align*}
 \begin{vmatrix}
      \cos{\alpha} & -\sin{\alpha} & 0 \\
      \sin{\alpha} & \cos{\alpha} & 0 \\
      0 & 0 & 1
 \end{vmatrix}.    
\end{align*}


\noindent$\alpha$ is the angle which governs whether the quadrotor will move right or left. {We limit $\alpha$ to the range $\left[{-90\degree}, {90\degree} \right]$.} For each angle of rotation $\alpha$ about the Z-axis, we again apply a rotation about the Y-axis as,
\begin{equation}
    v_{\alpha \beta} = R_{Y\beta} * v_{\alpha}
\end{equation}

where $\beta$ is the angle which governs whether the quadrotor will move upward or downward, and $\beta \in [-90\degree, 90\degree]$. The matrix $R_{Y\beta}$ is defined as,
\begin{align*}
    \begin{vmatrix}
      \cos{\beta} & 0 & \sin{\beta} \\
      0 & 1 & 0 \\
      -\sin{\beta} & 0 & \cos{\beta}
    \end{vmatrix}.    
\end{align*}

Let us denote the coverage area for $v_{\alpha \beta}$ over time $\tau$ as $area_{\alpha \beta}$. The rotation of $v_{unit}$ is continued until all values of $\alpha$, and $\beta$ within their ranges are reached by increments of 1$\degree$ , and $v_{\alpha \beta}$ and the overlap between $area_{\alpha \beta}$ and $area_{unit}$ are recorded into the LUT. The fully constructed LUT has 32,761 x 5 entries. The structure of the LUT is shown in Table \ref{tab:LUT}. 

\begin{table}
    \begin{tabular}{|c|c|c|c|c|}
      \hline
      \textbf{$\alpha$ (Yaw)} & \textbf{$\beta$ (pitch)} & \textbf{$v_{\alpha\beta}$} & \textbf{$v_{unit} - v_{\alpha\beta}$} & \textbf{$area_{unit} \cap area_{\alpha\beta}$} \\
      \hline
      & & & & \\
      \hline
    \end{tabular}
    \caption{\label{tab:LUT} The structure of the Look-Up Table.}
\end{table}

\subsubsection{Selecting New Velocity from LUT}
When the quadrotor has to change its velocity to avoid an obstacle, we first use ORCA to compute a new velocity $v_{new}$ as given in (\ref{Eqn.7}). The difference $\delta$ between $v_{new}$ and $v_{pref}$ is given as,
\begin{equation}
    \delta = |v_{pref} - v_{new}|.
\end{equation}
Next, we search the LUT for $v_{\alpha \beta}$ such that 
\begin{equation} \label{delta}
    |v_{unit} - v_{\alpha \beta}| \ge \delta + \varepsilon
\end{equation}
 where $\varepsilon$ is a small positive value. All the velocities  which satisfy $\left(\ref{delta}\right)$ are ranked based on their corresponding overlap area with $area_{unit}$. The Z-component of $v_{\alpha \beta}$ is used to compute the altitude change in the time horizon, and the velocities whose altitude change fail the GSD constraint are neglected. The rotation $R_{trans}$ is applied to each shortlisted $v_{\alpha \beta}$ to transform it correspond to the preferred velocity orientation. Let us denote the transformed velocity as $v_{\alpha \beta}^{trans}$. It is computed as:
\begin{equation}
    v_{\alpha \beta}^{trans} = R_{trans} \cdot v_{\alpha \beta}
\end{equation}

Note that there will always be a velocity that is equal to or closest to $v_{new}$ given by ORCA in the list of $v_{\alpha \beta}^{trans}$. We check if each $v_{\alpha \beta}^{trans} \in ORCA_{quadrotor}^{\tau}$ which is the ORCA set for the quadrotor for time $\tau$. The transformed velocity $v_{\alpha \beta}^{trans}$ with the highest rank that belongs to the ORCA set is guaranteed to avoid collisions and such a $v_{\alpha \beta}^{trans}$ may or may not be equal to the new velocity given by ORCA. Thus, the coverage area resulting from choosing this velocity will always be greater than or equal to the coverage area resulting from the velocity given by ORCA. \\  

\noindent {\bf{Computed Behaviour: }} ORCA computes a collision avoiding velocity for an agent considering the position and velocity of its neighbors. The collision avoiding velocity thus computed is the closest distance away from the preferred velocity. In contrast, LSwarm replaces the closest distance constraint in order to identify a velocity that gives better coverage overlap from the collision free velocity region. The LUT provides a pre-computed sample of all velocities and their coverage overlap to facilitate choosing the new velocity quickly rather than performing the overlap area computation in each iteration.


\begin{algorithm}
\caption{LSwarm}

\begin{algorithmic}[1]

\Procedure{LSwarm}{}
    \For{all quadrotors}
        \State Get position and velocity of Agent A
        \State Compute nearest neighbors set NN for A
        \For{all agent B in NN}
            \State Get position and velocity for B
            \State Compute Velocity Obstacle
            \State Construct ORCA Planes
        \EndFor
        \State Construct Collision Free Velocity Space
        \If{A lies in the optimal path = True}
          \State ${V_A}^{Pref} \gets normalize({WP_A}^{Next} - {P_A})$
        \Else
          \State ${V_A}^{Pref} \gets normalize({CP_A} - {P_A})$
        \EndIf
        \State Compute $V_A^{New}$ using ORCA
        \State Search LUT for velocities $\ge$ $V_A^{pref} - V_A^{New}$ 
        \State remove velocities with with fail GSD constraint
        \For{every velocity V in FV}
          \If{V satisfies ORCA planes}
            \State ${V_A}^{New} \gets V$
          \EndIf
        \EndFor
        \State Publish Velocity to Agent
    \EndFor
\EndProcedure

\end{algorithmic}
\end{algorithm}
\section{RESULTS AND PERFORMANCE}
In this section, we describe our implementation and highlight LSwarm's performance on different urban scenarios.

\begin{figure}[t]
    \centering
    \includegraphics[height=2.3in, width=3.5in]{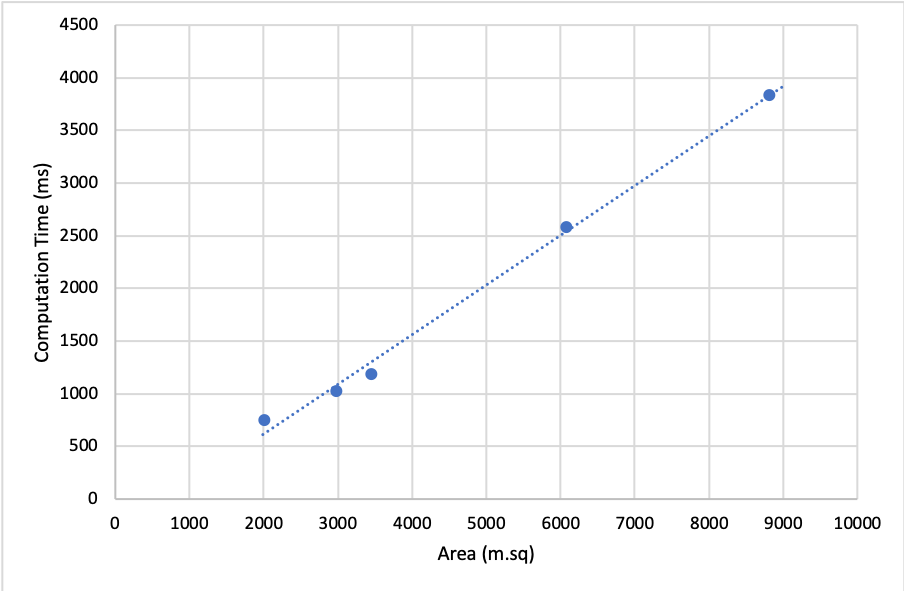}
    \caption{Variation of Lawnmower Computation time with Model Area. Since the computation time is linear, we conclude that the lawnmower strategy is scalable.}
    \label{fig:lawnmowerscalability}
\end{figure}

\begin{figure*}[t]
    \centering
    \begin{subfigure}[b]{0.3\textwidth}
        \includegraphics[width=\textwidth]{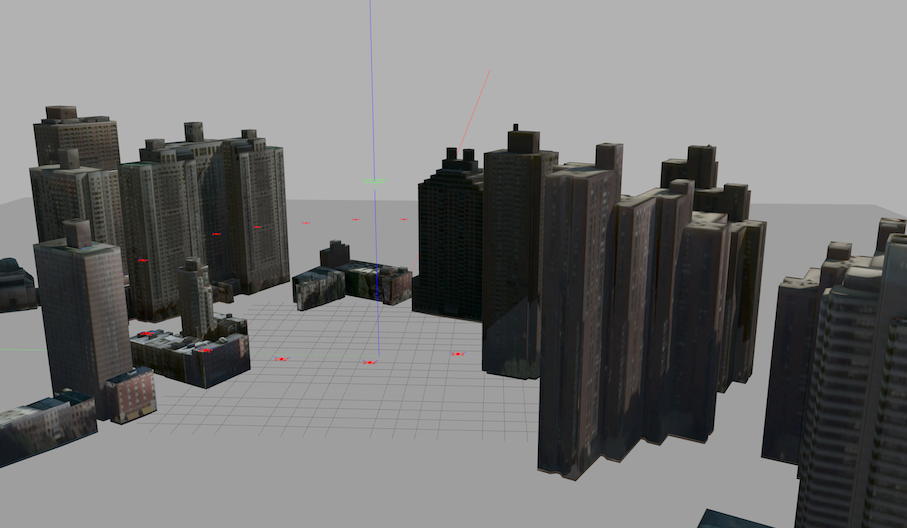}
        \caption{\textbf{High Sparse:} Sparsely populated urban model with 16 skyscrapers}
        \label{fig:highspare}
    \end{subfigure}
    ~ 
    \begin{subfigure}[b]{0.3\textwidth}
        \includegraphics[width=\textwidth]{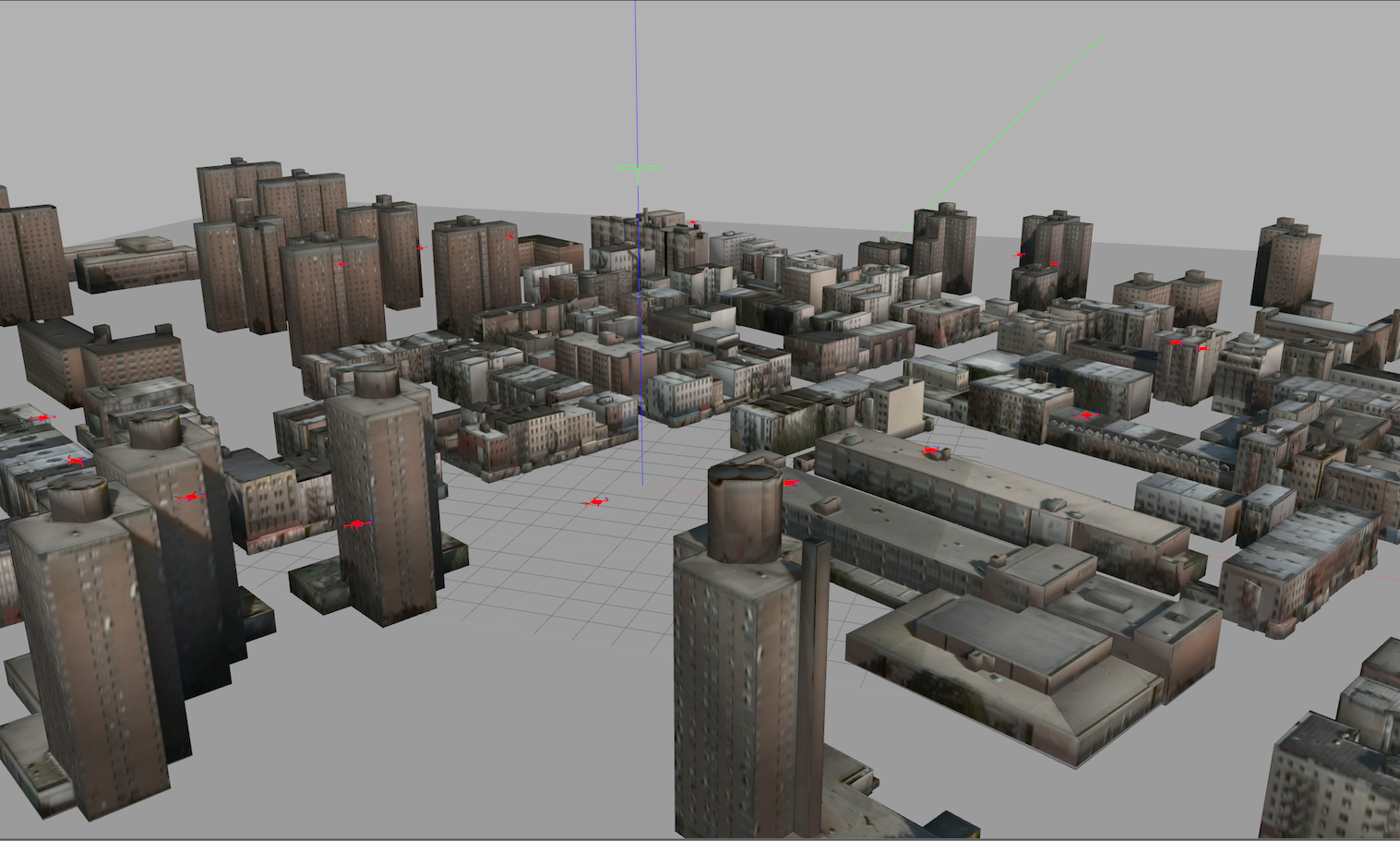}
        \caption{\textbf{Low Dense:} Sparsely populated model with 23 small buildings}
        \label{fig:lowsparse}
    \end{subfigure}
    ~ 
    \begin{subfigure}[b]{0.3\textwidth}
        \includegraphics[width=\textwidth]{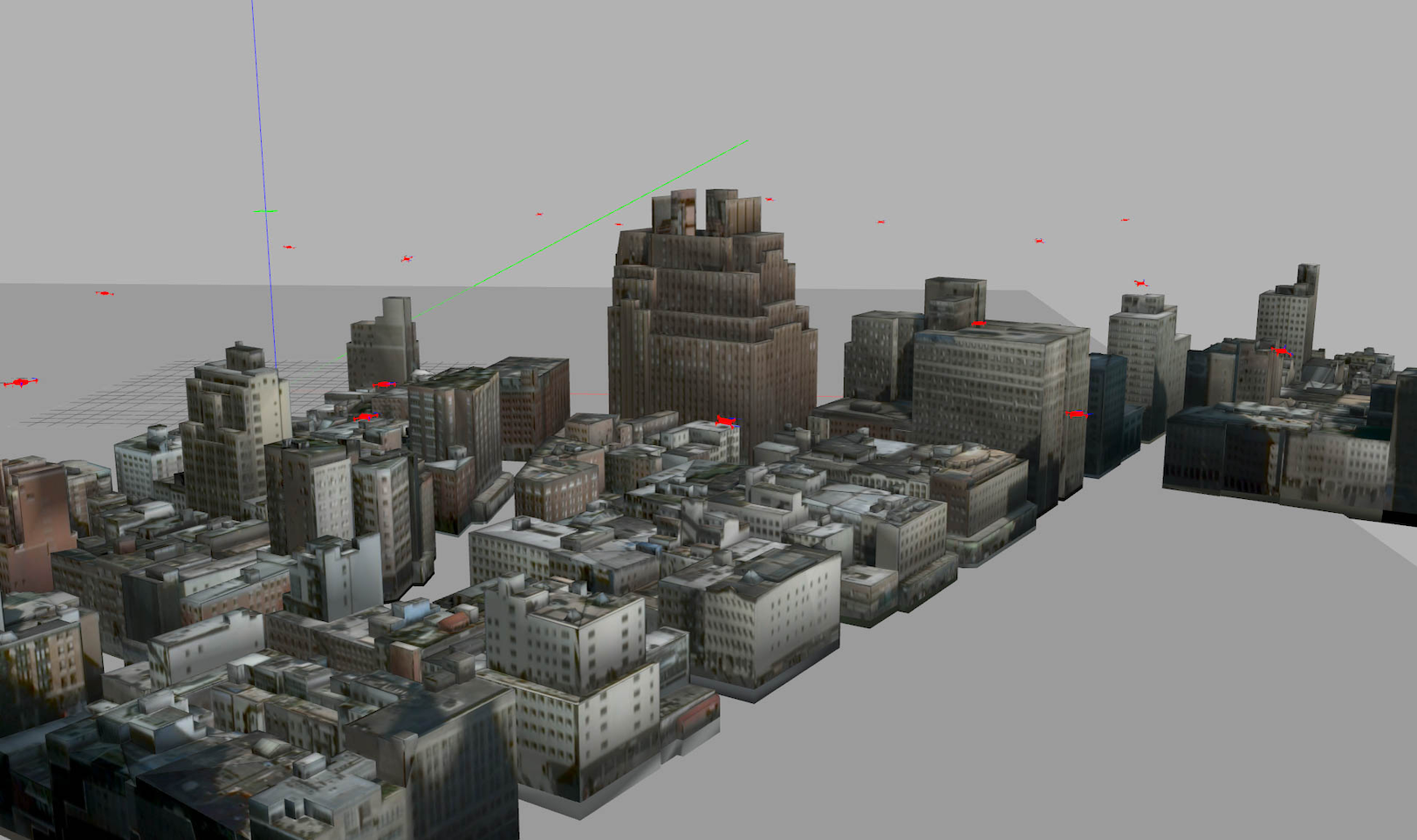}
        \caption{\textbf{Low Dense:} Densely packed model with 79 buildings of low height}
        \label{fig:lowdense}
    \end{subfigure}
    \caption{Benchmark urban environments}\label{fig:benchmark}
\end{figure*}

\subsection{Implementation}
LSwarm was implemented on an Intel Xeon w-2123 octacore processor at $3.6$ GHz with $32$ GB memory and GeForce GTX 1080 GPU. For simulating the swarm of quadrotors, we use ROS Kinetic, Gazebo 7.14.0, along with the PX4 Software In The Loop firmware. 

\subsubsection{LawnMower Strategy}
LSwarm can handle any arbitrary 3D model. X-Y values of the global waypoints are computed using an occupancy grid constructed from the 2D footprint of the environment, while the Z values vary based on the building heights (Section III.C). 

\subsubsection{Local Collision Avoidance}
ORCA is implemented using the RVO2-3D \cite{c29} library.  The RVO2-3D library makes use of multiple cores for decentralized collision avoidance among multiple agents. { In LSwarm, we compute the Euclidean separation distance for static collision avoidance using the Proximity Query Package (PQP) library \cite{c16}. We do not include any static obstacle avoidance as a part of ORCA in our comparisons. ORCA models static obstacles as line segments or planes, which is impractical for real world scenarios. Hence, we test ORCA with the lawnmower strategy (which accounts for static obstacles) to test if it would be adequate to avoid collision with buildings.}

\subsection{Benchmarks}
Table \ref{tab:Table 2} shows the dimensions and complexity of the different urban models used to evaluate our method, and the time taken to compute the global lawnmower waypoints. Three urban models are shown in Fig. \ref{fig:benchmark}. Fig.\ref{fig:highdense} shows the High-Dense urban model. The first word in the name of the models corresponds to the average height of the buildings (high or low) and the second word in the name signifies the density of buildings in the model.

The 4 benchmarks with varying building heights and building densities (buildings/sq.m) capture the real world scenarios that the aerial surveillance system may face. For example, the High Dense model (0.022 buildings/ meter.sq) represents the dense metropolitan scenario, and the Low Sparse model (0.003 buildings/meter.sq) represents rural or suburban situations.


\begin{table*}
\centering
\begin{tabular}{|c||c|c|c|c|c|}
 \hline
 \multirow{2}{*}{Environment Model} &\multirow{2}{*}{Dimension (in meters)} &\multirow{2}{*}{Number of Buildings} &\multirow{2}{*}{Global Path Calculation Time (ms)} &\multicolumn{2}{c|}{Collisions with buildings}\\
 \cline{5-6}
 &&&& ORCA & LSwarm \\
 \hline
 High Dense   & 50.96 x 39.33 x 29.50 &27 &753 &4 &nil\\
 High Sparse  & 56.25 x 53.03 x 14.25 &16 &1027 &2 &nil\\
 Low Dense    & 64.26 x 53.80 x 12.5  &79 &1186 &nil &nil\\
 Low Sparse   & 96.67 x 62.92 x 7.2   &23 &2585 &nil &nil\\
 \hline
\end{tabular}
\caption{\label{tab:Table 2} We present the complexity of different urban models used to evaluate our algorithm. Moveover, we highlight the time taken to compute the lawnmower waypoints and the number of collision with buildings when using ORCA and LSwarm.}
\end{table*}

\begin{table*}
\centering
\begin{tabular}{|c||p{2.5cm}|p{2.5cm}|p{2.5cm}|p{2.5cm}|}
 \hline
 \multirow{2}{*}{Number of Dynamic Obstacles} &\multicolumn{2}{c|}{ORCA} &\multicolumn{2}{c|}{LSwarm}\\
 \cline{2-5}
 & Overlap Ratio (with required Resolution) 1m Coverage Radius & Overlap Ratio (with required Resolution) 3m Coverage Radius & Overlap Ratio (with required Resolution) 1m Coverage Radius & Overlap Ratio (with required Resolution) 3m Coverage Radius \\
\hline
{\bf{Approaching from all directions}} &&&&\\
 10  & 0.7210 &0.9235 &0.8977 &0.9355\\
 25  & 0.6594 &0.7040 &0.8715 &0.9197\\
 40  & 0.5916 &0.6331 &0.6084 &0.8875\\
{\bf{Approaching from Left to Right}} &&&&\\
 10  & 0.7762 &0.8879 &0.9308 &0.9448\\
 20  & 0.3201 &0.5374 &0.8923 &0.9379\\

 \hline
\end{tabular}
\caption{\label{tab:Table 3} We present the complexity of different urban models used to evaluate our algorithm. Moreover, we highlight the time taken to compute the lawnmower waypoints using our global solution.}
\end{table*}

\subsection{Performance Evaluation}

\subsubsection{Lawnmower Path Generation}
The time taken to compute waypoints is proportional to the model size (i.e. area). For a model of area $\sim$ $6000$ $m^2$, the lawnmower strategy took $2.5$ seconds to generate the waypoints. From Fig. \ref{fig:lawnmowerscalability}, we infer that the lawnmower strategy can easily be scaled to larger urban scenes. In our implementation, the lawnmower strategy just acts as a placeholder. Any global coverage strategy that provides a set of waypoints for each agent, and has similar computation time as the lawnmower strategy, can be used instead.  

\begin{figure}[t]
      \centering
      \includegraphics[height=2.0in, width=3.5in]{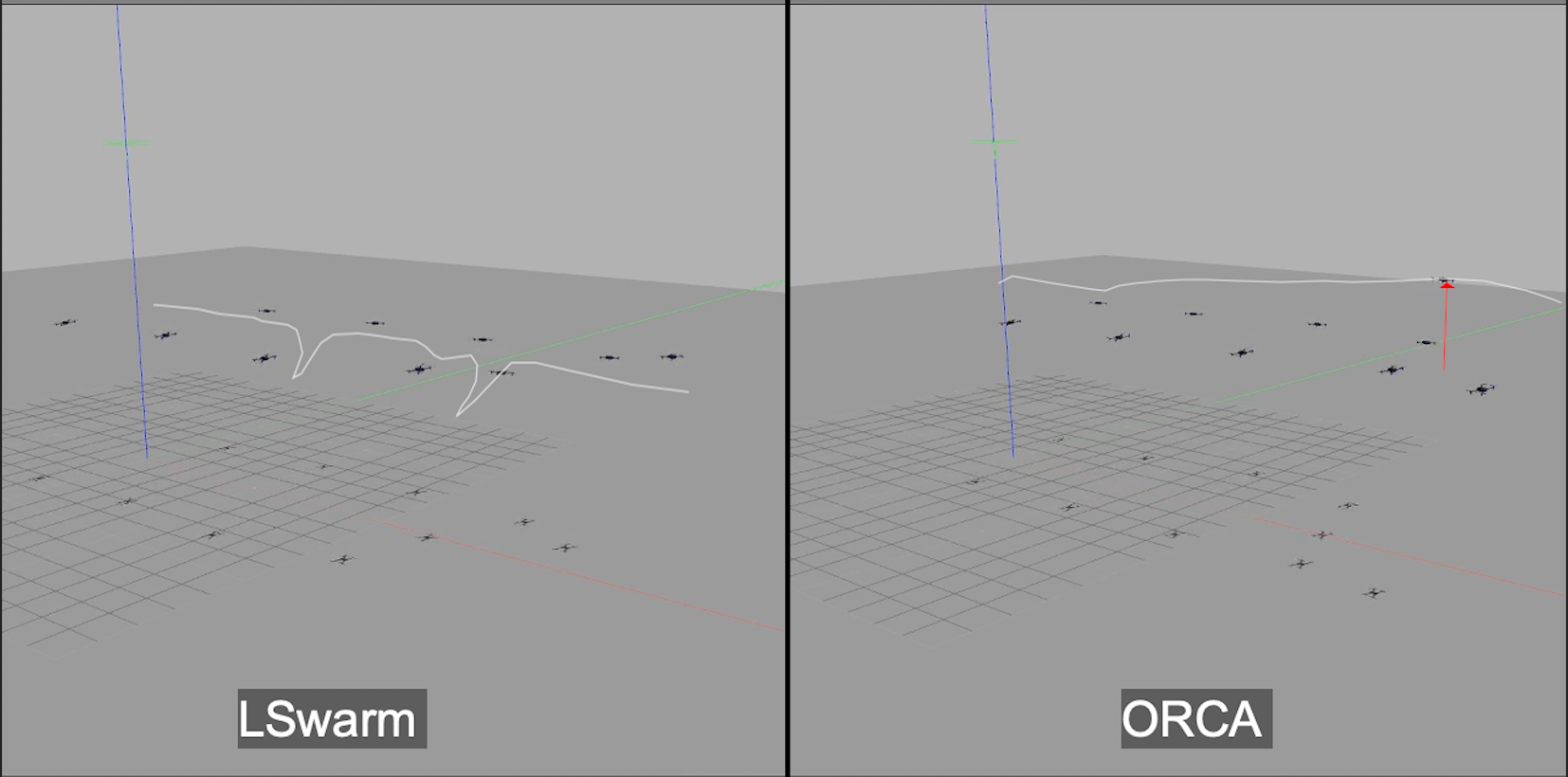}
      \caption{Trajectories followed by agents using LSwarm and ORCA in a scenario with 10 dynamic obstacles moving perpendicular to the motion of the agent. The red arrow shows the increase in an agent's altitude when ORCA was used for dynamic obstacle avoidance.}
      \label{fig:Traj}
   \end{figure}

{ \subsubsection{Comparing trajectories}
In the scenario shown in Fig.\ref{fig:Traj}, the trajectory given by LSwarm is less smooth than the one provided by ORCA. The agent using ORCA takes a smooth trajectory above all obstacles, which leads to poor unusable resolutions. However, the agents using LSwarm is forced to satisfy the resolution and coverage area constraints, leading it to maintain its altitude and fly through the dynamic obstacles.  Therefore, it is forced to make more deviations to avoid collisions. 

The smoothness of the trajectory given by LSwarm is scenario-dependent. In cases with fewer dynamic obstacles, the smoothness of LSwarm converges to that of ORCA. For instance, in the scenario shown in Fig.\ref{staticcollisionavoidance}, with each agent avoiding just a single dynamic obstacle and a static obstacle, LSwarm deviates agents much less than ORCA, and the trajectories have comparable smoothness to ORCA’s trajectories.}

\subsubsection{Local Collision Avoidance}
In this section, we compare the static collision avoidance and coverage performance of LSwarm and ORCA. { LSwarm provides the same agent-agent collision avoidance guarantees as ORCA and hence results pertaining to this case are not included in this paper.}

We evaluated the combined strategy in the 4 benchmark models, using 20 quadrotors and 20 dynamic obstacles. From Table \ref{tab:Table 2}, we observe that with higher building density and taller buildings, a portion of the quadrotor fleet collided with the buildings when ORCA was used. In contrast, LSwarm was effective in avoiding collisions with buildings. 

For coverage performance, we tested LSwarm by making the quadrotor cover a 20m straight line with obstacles approaching from different directions. We tested for scenarios when obstacles approach from all directions, approach perpendicular to the path direction, and approach from various heights. We observe that the improvement provided by LSwarm is prominent when the camera's coverage radius is small. For ensuring required resolution, we try to avoid increasing the quadrotor altitude by over 1.5 meters. Table \ref{tab:Table 3} summarizes the overlap ratio (considering constraints on resolution) for both ORCA and LSwarm. 

The coverage performance of LSwarm with respect to ORCA for a varying number of dynamic obstacles is shown in Fig. \ref{fig:effectiveoverlap}. For a small number of dynamic obstacles, there is little difference between the two methods. But as the number of dynamic obstacles increases, ORCA computes new velocities that may lead to coverage data with poor, unusable resolutions that cannot be considered. LSwarm consistently produces velocities that lead to nearly optimal coverage with useful resolutions.

\begin{figure}[t]
      \centering
      \includegraphics[height=2.3in, width=3.5in]{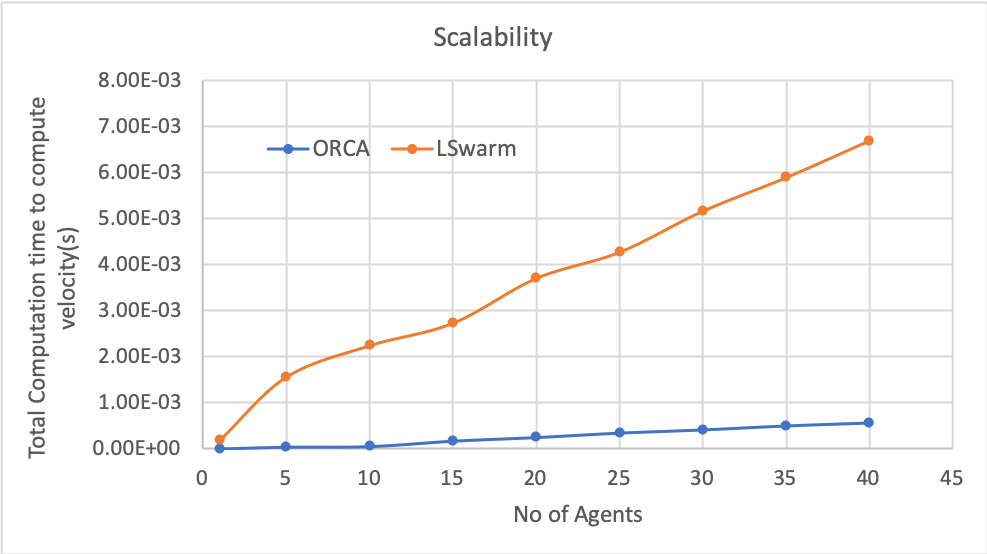}
      \caption{ Running time of LSwarm algorithm and ORCA vs. number of agents on a multi CPU-core. We observe nearly linear growth for LSwarm indicating scalability with the number of agents. As expected, ORCA takes less time to compute collision avoidance velocities.}
      \label{fig:lswarmscalability}
   \end{figure}

\begin{figure}[t]
      \centering
      \includegraphics[height=2in,width=3.25in]{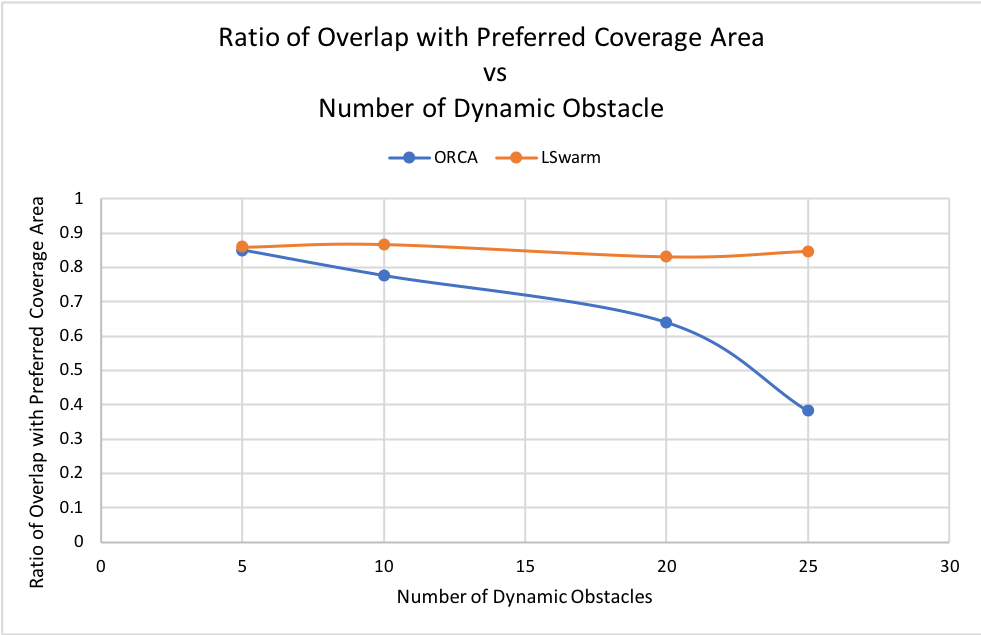}
      \caption{This graph shows the ratio of overlap area with the preferred area vs the number of dynamic obstacles while considering GSD constraints. LSwarm provides a better overlap for a large number of dynamic obstacles over ORCA.}
      \label{fig:effectiveoverlap}
   \end{figure}

The nearly linear relationship between time for velocity calculation in LSwarm and the number of swarm agents is shown in Fig. \ref{fig:lswarmscalability}. We infer that LSwarm can be scaled up with respect to the number of swarm agents. We also observe that LSwarm takes nearly 7 times more time for computation when compared with ORCA. This is due to the additional time taken for proximity queries when the PQP library is used. \\

\section{CONCLUSION, LIMITATIONS, AND FUTURE WORK}
We present an efficient method for local collision avoidance under coverage constraints for a large number of agents in a swarm. We pose the problem using an optimization formulation based on coverage constraints. In order to solve the non-linear optimization problem, we use a precomputed lookup table to compute a solution that provides guarantees on collision-free trajectories as well as the coverage area. The running time of our method scales linearly with the number of swarm agents and takes a few milliseconds for tens of swarms covering a large urban scene.      

Our approach has some limitations.{ Currently, we use a simple Kalman filter to address the uncertainties in localization of agents/obstacles whereas uncertainty in actuation is not considered.} Furthermore, the lawnmower algorithm assumes an exact representation of the urban environment. Our solution to the optimization formulation is approximate and does not provide guarantees with respect to global optimality.

There are many avenues for future work. In addition to addressing these limitations, we would like to further evaluate our approach on complex urban models by coupling with other coverage strategies. It would be useful to develop more robust methods that can handle noise in the sensor measurements and account for uncertainty. Finally, we would like to accurately model the dynamics constraints of swarm agents (e.g. quadrotors).

\end{document}